%% file: egpaper_for_review.tex
\documentclass[10pt,twocolumn,letterpaper]{article}

\usepackage{iccv}
\usepackage{times}
\usepackage{epsfig}
\usepackage{graphicx}
\usepackage{amsmath}
\usepackage{amssymb}
\usepackage{amsthm}
\usepackage{bm}
\usepackage{makecell}
\usepackage{subfigure}
\usepackage{booktabs}
\usepackage{authblk}
\newcommand*{\Scale}[2][4]{\scalebox{#1}{$#2$}}%

\usepackage[pagebackref=true,breaklinks=true,letterpaper=true,colorlinks,bookmarks=false]{hyperref}

\iccvfinalcopy % *** Uncomment this line for the final submission

\def\iccvPaperID{6738} % *** Enter the ICCV Paper ID here

\ificcvfinal\fi
\newtheorem{prop}{Proposition}
\begin{document}

%%%%%%%%% TITLE

\title{AdaFit: Rethinking Learning-based Normal Estimation on Point Clouds}

\author{
Runsong Zhu$^1$\thanks{Equal contribution} \quad
Yuan Liu$^2$$^*$ \quad
Zhen Dong$^1$\thanks{Correspondig authors: [dongzhenwhu,bshyang]@whu.edu.cn} \quad 
Tengping Jiang$^1$ \quad 
Yuan Wang$^1$ \quad 
Wenping Wang$^{2,3}$ \quad 
Bisheng Yang$^1$$^\dagger$ \\ 
$^1$Wuhan University \quad $^2$The University of Hong Kong \quad $^3$Texas A\&M University
}

\maketitle

%%%%%%%%% ABSTRACT

%%%%%%%%% ABSTRACT

\input{00_abstract}
\input{01_introduction}

\input{02_related_works}
\input{03_method}

\input{04_experiments}
\input{05_applications}
\input{06_conclusion}
\input{08_Acknowledgement}

% \input{ada-Net/LaTeX/07_supplementary}

{\small
\bibliographystyle{unsrt}
\bibliography{egpaper_for_review}
}

\end{document}

%% file: 00_abstract.tex
\begin{abstract}

This paper presents a neural network for robust normal estimation on point clouds, named AdaFit, that can deal with point clouds with noise and density variations. Existing works use a network to learn point-wise weights for weighted least squares surface fitting to estimate the normals, which has difficulty in finding accurate normals in complex regions or containing noisy points. By analyzing the step of weighted least squares surface fitting, we find that it is hard to determine the polynomial order of the fitting surface and the fitting surface is sensitive to outliers. To address these problems, we propose a simple yet effective solution that adds an additional offset prediction to improve the quality of normal estimation. Furthermore, in order to take advantage of points from different neighborhood sizes, a novel Cascaded Scale Aggregation layer is proposed to help the network predict more accurate point-wise offsets and weights. Extensive experiments demonstrate that AdaFit achieves state-of-the-art performance on both the synthetic PCPNet dataset and the real-word SceneNN dataset. The code is publicly available at \href{https://github.com/Runsong123/AdaFit/}{https://github.com/Runsong123/AdaFit}.
\end{abstract}

%% file: 01_introduction.tex
\section{Introduction}

In point cloud processing, a fundamental task is to robustly estimate surface normals from point clouds, which plays a key role in many practical applications, such as surface reconstruction~\cite{kazhdan2006poisson}, registration~\cite{pomerleau2015review}, segmentation~\cite{grilli2017review}, primitive fitting~\cite{schnabel2007efficient}, reverse engineering~\cite{liu2020adaptive} and grasping~\cite{zapata2019fast}. Due to the presence of noise, point density variations, and missing structures, robust and accurate surface normal estimation on point clouds remains challenging.

\begin{figure}
   \begin{center}
      \includegraphics[width=0.95\linewidth]{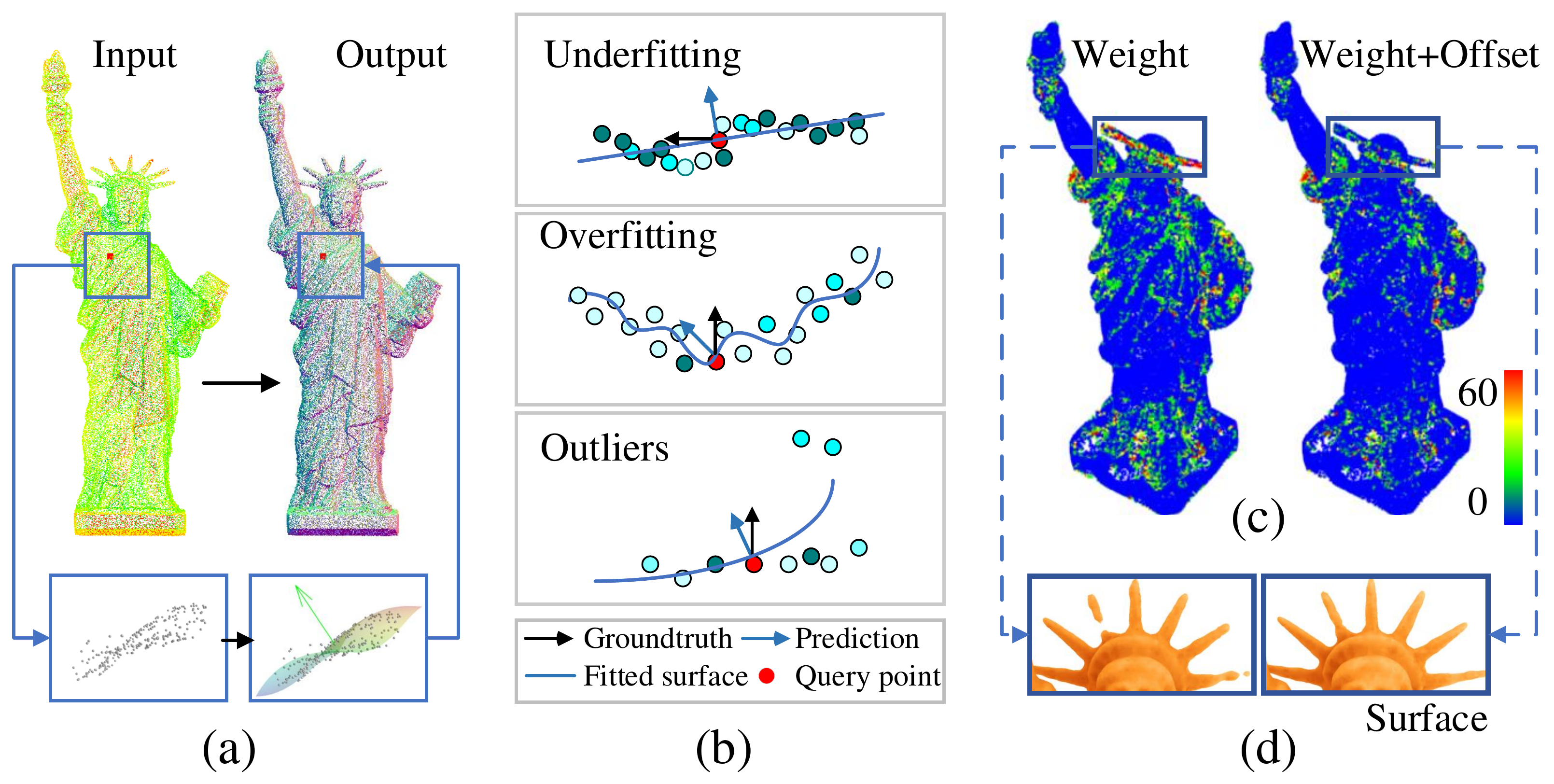}\vspace{-0.4cm}
   \end{center}
   \caption{(a) Given the input point cloud (left), our target is to estimate a normal for every point (right). (b) Current weighted least square surface fitting is severely affected by underfitting (top), overfitting (middle) or outliers (bottom), which leads to inaccurate normal estimation. (c) The error maps of two normal estimation methods. The first model only uses the weighted least square surface fitting (left) while the second model adds additional offsets to adjust the distribution of neighboring points, which produces a more accurate normal estimation. (d) The reconstructed surfaces using normals from the model with weights only (left) and the model with weights and offsets (right).}
   \label{fig:intro}
   \vspace{-0.25cm}
\end{figure}

The most direct way for normal estimation is to regress the normal vector from the feature extracted on neighboring points~\cite{guerrero2018pcpnet,ben2019nesti,zhou2020normal,hashimoto2019normal,pistilli2020point,zhou2020geometry}. However, such brute-force regression only forces the network to memorize normal vectors, which leads to limited generalization ability. This generalization problem becomes more severe on real-world data due to the scarcity of training data.

Rather than direct regression, a more accurate approach to estimate the normal for a specific point is to fit a geometric surface (plane or polynomial surface) on its neighboring points and then compute the normal from the estimated surface. Since the estimated surface is usually sensitive to the noise or outliers present in the neighborhood, most existing methods~\cite{ben2020deepfit,holland1977robust,lenssen2020deep} use weighted surface fitting which predicts point-wise weights to control the contribution of every neighboring point to the final surface. The focus of these works is to obtain more accurate point-wise weights that can reduce the effects of noisy points and outliers as possible. Though substantial improvements on normal estimation have been achieved by using a neural network to learn a more accurate weight~\cite{ben2020deepfit,lenssen2020deep} in a data-driven manner, the estimated normals are still not accurate in complex regions as shown in Fig.~\ref{fig:intro} (c).

In this paper, we conduct an analysis on the widely-used weighted surface fitting for normal estimation and find two inherent problems in this approach. The first is brought by the inconsistent polynomial orders between the true surface and the fitted surface. The underlying surfaces of different points usually have different polynomial orders while current methods always choose a constant order for all points, e.g. plane in~\cite{lenssen2020deep} or 3-jet in~\cite{ben2020deepfit}. Such inconsistency either results in an underfitting, which smooths out the fine details in the neighborhood shown in the top of Fig.~\ref{fig:intro} (b), or overfitting to noise, which brings a large variance to the output normal shown in the middle of Fig.~\ref{fig:intro} (b). Both the overfitting and underfitting may result in an erroneous normal estimation. The other problem is that the weighted surface fitting is sensitive to outliers. By theoretically analyzing the relationship between point-wise weights and the final estimated normal, we find that the weight on a point far away from the fitted surface will have a larger impact on the final normal direction. In this case, since outliers are much far away from the fitted surface than inlier points, even small weights on outliers will thoroughly mess up the estimated normals, as shown in the bottom of Fig.~\ref{fig:intro} (b).

To address the above two problems, we propose a simple yet effective solution by predicting additional point-wise offsets to adjust the distribution of the neighboring points. When the underlying surface has a different polynomial order from the predefined one, the adjustment brings more flexibility to the network to project neighboring points onto a surface with the predefined order. Though the resulted surface may not be precisely consistent with the ground truth one, the normal of the center point is much more accurate than the one estimated by the direct weighted surface fitting. Meanwhile, the outliers in the neighborhood can be offset to a position near the resulted surface such that all points will have similar distances to the surface. Thus, in comparison with the weighted surface fitting where the weights of outliers have larger impacts on the normal, points contribute more evenly after the adjustment, which leads to a more robust normal estimation.

Another challenge is the selection of optimal neighborhood size in the surface fitting. Here, we call the neighborhood size as scale. A large scale containing more points will provide more information about the underlying surface but may include irrelevant points which easily leads to over-smoothing sharp edges. A small scale only contains a small set of most relevant points thus may improve the accuracy of the normal estimation but is unavoidably sensitive to noise. The scale is a very sensitive hyperparameter that requires a carefully tuning for an accurate normal estimation~\cite{ben2020deepfit,lenssen2020deep}.

To address the problem of the scale selection, we use a novel architecture design called Cascaded Scale Aggregation (CSA) layer. With several CSA layers, we can extract features from the large scale while only fit the surface with neighboring points in a small scale. Thus, using CSA layer for feature extraction enjoys the benefits from both the large scale, which brings more information about the surface, and the small scale, which results in a more accurate normal estimation.

To this end, we implement our idea by designing a network called AdaFit, which takes a query point with its neighboring points as input and outputs the normal of the query point. AdaFit extracts features with CSA layers and simultaneously predicts point-wise weights and offsets. Then, the predicted weights are used to fit a polynomial surface of order 3 on these offset neighboring points. Finally, the output normal is computed from the fitted surface. 

To validate the proposed method, we conduct extensive experiments on the widely-used PCPNet dataset~\cite{guerrero2018pcpnet}. The results show that the proposed AdaFit achieves state-of-the-art performance on this benchmark. To show the generalization ability of our AdaFit, we evaluate it on two real-world datasets, the indoor SceneNN~\cite{hua2016scenenn} dataset and the outdoor Semantic 3D~\cite{hackel2017isprs} dataset. Without any further training, AdaFit significantly outperforms baselines by a large margin on these two datasets. Furthermore, we demonstrate the applications of the normals predicted by AdaFit on the point cloud denoising and the surface reconstruction.

Our contributions are summarized as follows:
\begin{itemize}
    \item We provide a comprehensive analysis on the weighted surface fitting and find two critical problems of these methods in normal estimation.
    \item We propose to predict offsets to adjust the distribution of neighboring points which brings more robustness and accuracy in normal estimation.
    \item We design the network AdaFit with novel CSA layers to enjoy benefits from both small and large scales, which achieves improved performance in multiple standard datasets.
   
\end{itemize}
%-------------------------------------------------------------------------

%------------------------------------------------------------------------

%% file: 02_related_works.tex
\section{Related work}

\subsection{Traditional normal estimation}

The most popular method for normal estimation is based on Principal Component Analysis (PCA)~\cite{hoppe1992surface} and Singular Value Decomposition (SVD)~\cite{klasing2009comparison}, which are utilized to find the eigenvectors of the covariance matrix constructed from the neighboring points. These approaches heavily depend on the selected scale and are sensitive to noise and outliers. Following this work, moving least squares (MLS)~\cite{levin1998approximation}, variants fitting local spherical surfaces~\cite{guennebaud2007algebraic}, Jets~\cite{cazals2005estimating} (truncated Taylor expansion) are proposed to fit more complex local surface. Generally, they choose a large-scale neighborhood size to improve the robustness, but tend to the oversmooth details. Mitra \etal~\cite{mitra2003estimating} reduced the neighborhood size by finding a optimal radius $r$ from point density or curvatures of underlying surfaces.
%or curvatures of underlying surfaces.
To retain more detailed shape, several methods~\cite{merigot2010voronoi,amenta1999surface,alliez2007voronoi,boulchfast} utilize Voronoi cells or Hough transform. 
Although these techniques come with strong theoretical approximation and robustness guarantees, such methods need to manually set the parameters according to the noise levels in the raw point clouds.

\subsection{Learning-based normal estimation}

\textbf{Regression based methods}. With the success of deep learning in a wide range of domains~\cite{qi2017pointnet,qi2017pointnet++,zhang2020pointfilter,rakotosaona2020pointcleannet,hermosilla2019total,yu2018ec}, some deep learning-based normal estimation approaches are proposed, which utilize the powerful feature extraction capability of deep learning and transform the normal estimation task into a regression or classification task. Depending on the formats of input, learning-based methods can be divided into two groups. The first group of methods~\cite{boulch2016deep,zhou2020geometry,lu2020deep} transform unstructured point clouds into structured grid format and apply Convolutional Neural Network (CNN) for feature learning. For example, Boulch \etal~\cite{boulch2016deep} associate a 2D grid representation to the local neighborhood of a 3D point via a Hough transform, and formulate the normal estimation as a discrete classification problem in the Hough space. The second group of methods~\cite{guerrero2018pcpnet,ben2019nesti,zhou2020normal,hashimoto2019normal,pistilli2020point} directly estimate surface normals from unstructured point clouds. For instance, PCPNet~\cite{guerrero2018pcpnet} estimates surface normal by a deep multi-scale PointNet~\cite{qi2017pointnet} architecture, which processes the multiple neighborhood scales jointly, thus leading to the phenomenon of oversmoothing. To overcome the oversmoothing phenomenon, Nesti-Net~\cite{ben2019nesti} applied a MoE~\cite{jacobs1991adaptive} structure to predict the optimal scale rather than direct concatenation of multiple scales, which yields an improved performance. Similarly, Zhou \etal~\cite{zhou2020normal} improve surface normal estimation by using an extra feature constraint mechanism and a novel multi-scale neighborhood selection strategy. Hashimoto \etal~\cite{hashimoto2019normal} propose a joint model that exploits a PointNet for local feature extraction and a 3DCNN for the spatial feature encoding to efficiently incorporate local and spatial structures.

\textbf{Surface fitting based methods}.
The former learning-based normal estimation studies~\cite{guerrero2018pcpnet,ben2019nesti,zhou2020normal,hashimoto2019normal,pistilli2020point,zhou2020geometry} directly regress the surface normal with fully connected layers, which leads to weak generalization ability and unstable prediction results. To solve the these limitations, Lenssen \etal~\cite{lenssen2020deep} and DeepFit~\cite{ben2020deepfit} first utilize the network to predict point-wise weights acting as a soft selection of the neighboring points, then estimate the surface normal by the differentiable and weighted least squares plane or polynomial surface fitting. These methods heavily constrain the space of solutions to be better suited for the given problem and enable the computation of additional geometric properties such as principal curvatures and principal directions.
Our method also belongs to this category and we add additional offsets to make the output normals more robust and accurate. Moreover, we propose a novel CSA layer to aggregate features from multiple neighborhood sizes.

%% file: 03_method.tex
\section{Method}
\subsection{Problem statement}
Given a point $p$ and its neighboring points $\{p_i|i=1,...,N_p\}$, we want to estimate the normal $\mathbf{n}_{p}$ at the point. The normal estimation problem can be solved by fitting a surface on the neighboring points and compute the normal from the fitted surface. Here, we adopt a widely-used n-jet surface model~\cite{cazals2005estimating}, which represents the surface by a polynomial function $J_{n}:\mathbb{R}^2\to \mathbb{R}$ mapping coordinates $(x,y)$ to their height $z$ in the tangent space by
\begin{equation}
\Scale[1.00]{z=J_n(x,y;\hat{\beta})=\sum_{k=0}^{n} \sum_{j=0}^{k} \hat{\beta}_{k-j, j} x^{k-j} y^{j}},
    \label{eq:surface}
\end{equation}
where $\Scale[0.94]{\hat{\beta}}$ are coefficients. To simplify the notations, we denote the vector
$$\beta = \Scale[0.94]{\big(\hat{\beta}_{0,0}, \cdots,\hat{\beta}_{k,0},\hat{\beta}_{k-1,1},\cdots,\hat{\beta}_{1,k-1},\hat{\beta}_{0,k},\cdots\big)}$$ and we define $\beta_1 = \hat{\beta}_{1,0}$ and $\beta_2 = \hat{\beta}_{0,1}$.

Once the surface is fitted, the normal can be computed by:
\begin{equation}
    \mathbf{n}_{p} = \frac{(-\beta_{1},-\beta_{2},1)}{\sqrt{\beta_{1}^2+\beta_{2}^2+1}},
\end{equation}

In order to find the correct surface function in Eq.~\ref{eq:surface}, a point-wise weight is predicted on every point. Then, all points are transformed to the tangent space by Principle Component Analysis (PCA) and we solve for the surface coefficients by a weighted least square (WLS) fitting problem as follows
\begin{equation}
    \beta = \mathop{\rm argmin}_\alpha \sum_{i}^{N_p} w_i \|J_n(\alpha,x_i,y_i)-z_i\|^2,
    \label{eq:wls}
\end{equation}
where $w_i$ is the point-wise weight and $(x_i,y_i,z_i)$ is the coordinate of $p_i$ in the tangent space. The solution to Eq.~\ref{eq:wls} is given by   
\begin{equation}
    \beta = (M^\intercal W M)^{-1}(M^\intercal W z),
\end{equation}
where $W\in \mathbb{R}^{N_p\times N_p}$ is the diagonal matrix consisting of the $w_i$, $M\in \mathbb{R}^{N_p\times N_n}$ whose $i$-th row vector $M_i$ is $(1,x_i,y_i,...,x_i^2y_i^{n-2},x_{i}y_i^{n-1},x_i^n,y_i^n)$ and $z=(z_1,...,z_{N_p})\in \mathbb{R}^{N_p}$.

\subsection{Analysis on WLS for normal estimation}
To utilize the WLS for the normal estimation, most existing works focus on learning a more accurate weight for every neighboring point. However, two inherent problems of WLS prevent these methods from estimating a more accurate normal.

\begin{figure}[]
   \begin{center}
      \includegraphics[width=0.9\linewidth]{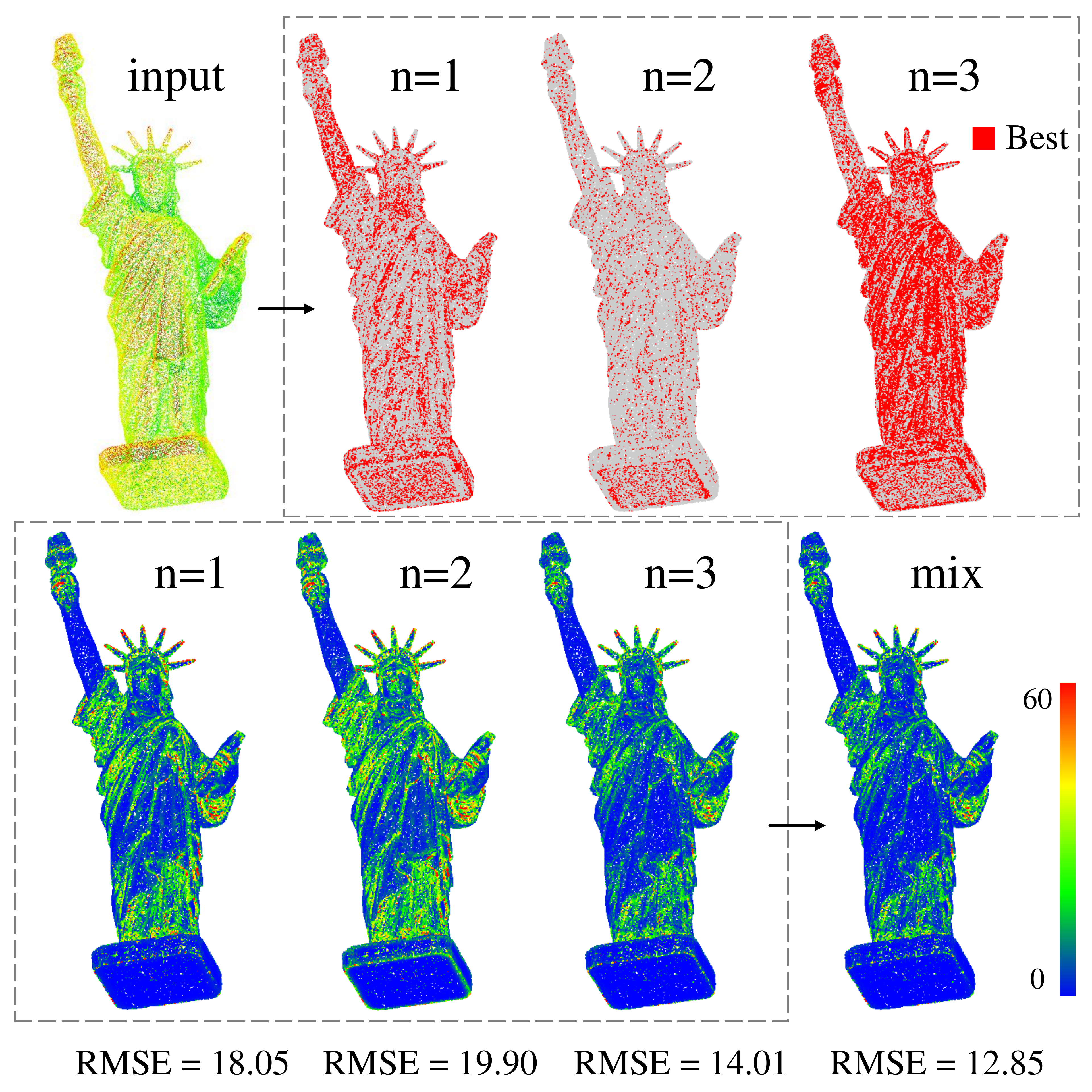}\vspace{-0.45cm}
   \end{center}
   \caption{The first row shows the best polynomial orders for different points. The second row shows the corresponding error map of different polynomial orders and the last one (``mix") always uses the best orders for different points for the normal estimation.}
   \label{fig:order}
  \vspace{-0.3cm}
\end{figure}

\textbf{Inconsistent polynomial orders}. For different points, their neighboring points usually conform to surfaces of different polynomial orders. However, the order $n$ in Eq.~\ref{eq:surface} is a constant predefined integer in WLS models for all points. The inconsistency of orders either brings overfitting or underfitting on the surface fitting. When the predefined order $n$ is smaller than the order of the ground truth surface, an underfitting occurs so that the model struggles to find a true normal for the point. Meanwhile, when the predefined order $n$ is larger than the true order, overfitting would make the fitting process sensitive to the noise on the neighboring points, thus brings instability on the estimated normals. A typical example is shown in Fig.~\ref{fig:order}, where we fit surfaces with different polynomial orders and red points in the first row show that different points have different suitable polynomial orders for the normal estimation. In the second row, we show the normal errors using different polynomial orders for surface fitting while in the last figure, we always choose their the best polynomial order for different points, which substantially reduces the overall normal RMSE to 12.85$^\circ$.

\textbf{Sensitivity to outliers}. Meanwhile, the WLS is also sensitive to outliers in the neighborhood. By checking how the weight of every neighboring point affects the final fitted surface, we can prove the following proposition.

\begin{prop}
    For a specific point $p_i$, if it is farther from the fitted surface in Eq.~\ref{eq:surface}, which means the predicted height $z_i'=J_n(\beta,x_i,y_i)$ on this point is largely deviated from the input height $z_i$, then the weight on this point will have a larger impact on the fitted surface, i.e. $\partial \beta / \partial w_i = (M^\intercal WM)^{-1}M_i^\intercal (z_i-z_i^{'})$. 
    \label{prop}
\end{prop}

In general, outliers will locate far away from the fitted surface with larger $z_i-z_i'$, the resulted surface coefficients $\beta$ will be more sensitive to their weights according to the Proposition~\ref{prop}. Moreover, we can directly compute the derivative of the estimated normal $\mathbf{n}_{p}$ to the point-wise weight $w_i$ by $\frac{\partial \mathbf{n}_{p}}{\partial w_i}=\frac{\partial \mathbf{n}_{p}}{\partial\beta}\frac{\partial \beta}{\partial w_i}$. Thus, even though the network may learn to place small weights on these outliers, a small perturbation on the weights of outliers will still lead to a significant change in the output normal.

\subsection{Offset prediction}
\begin{figure}[]
   \begin{center}
      \includegraphics[width=0.9\linewidth]{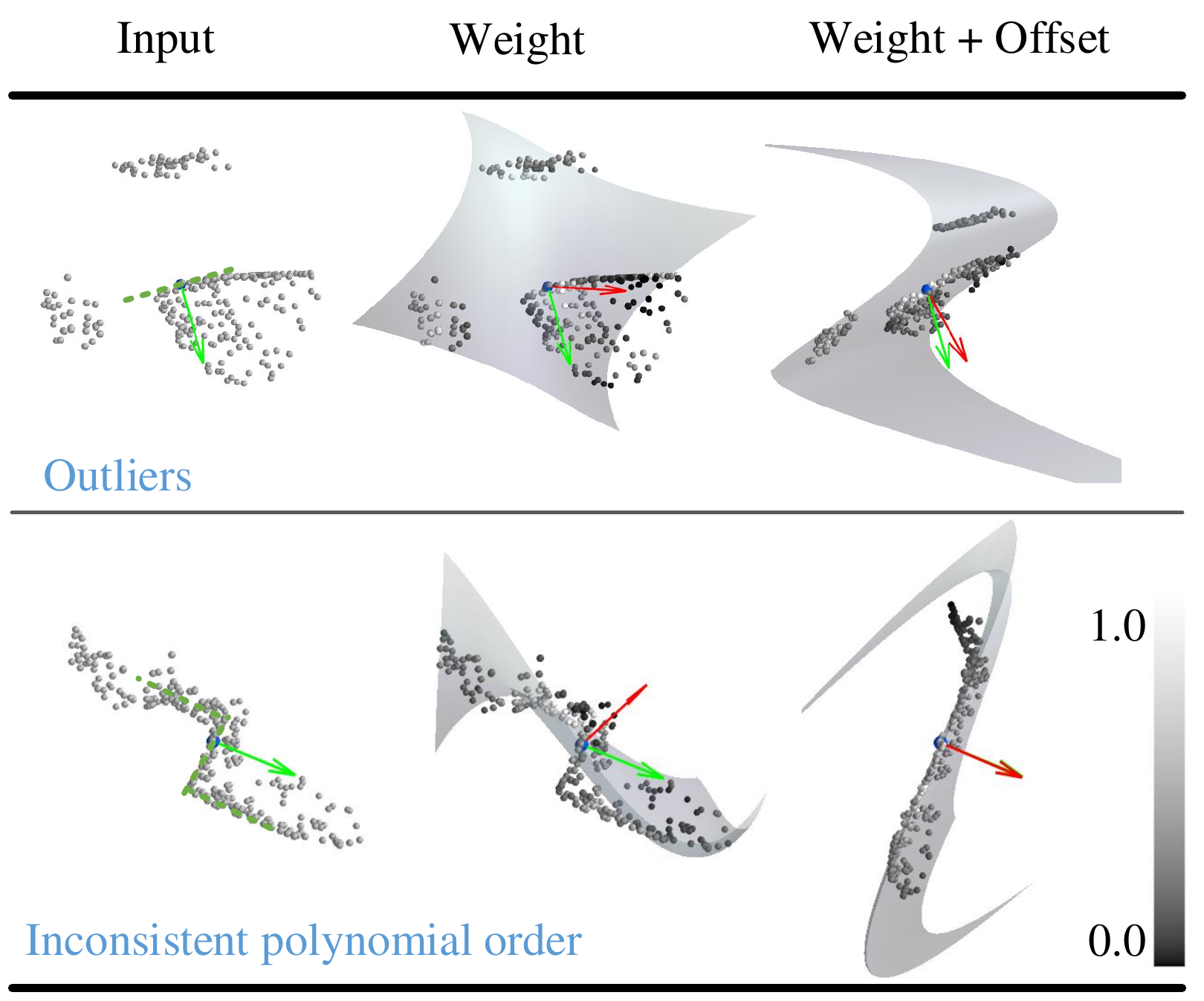}\vspace{-0.4cm}
   \end{center}
   \caption{Two examples of direct fitting the surface with weights and fitting with offset points. Green arrows are true normal directions while red arrows show the predicted normal directions. The first example contains some irrelevant outliers while the second one shows the underlying surface is too complex for the current polynomial order to achieve a good fitting. The offsets adjust the distribution of points, which produce a more accurate normal estimation in both cases.}
   \label{fig:fit_offset_compre}
   \vspace{-0.1cm}
\end{figure}

To address the above two problems, we propose a simple yet effective solution, in which we first predict an additional point-wise offset $(\Delta x_i, \Delta y_i, \Delta z_i)$ on every point to adjust the distribution of points. Then, WLS is applied on these offset points to find the surface by
\begin{equation}
    \beta = \mathop{\rm argmin}_\alpha \sum_i^{N_p} w_i \|J_n(\alpha,x_i+\Delta x_i,y_i+ \Delta y_i)-(z_i+\Delta z_i)\|^2.
\end{equation}
Due to the offset prediction, the network has an additional flexibility to adjust points to construct a virtual surface that has the same polynomial order as the predefined one. Thus, the offset prediction greatly reduces the underfitting or overfitting phenomenons. Meanwhile, outliers can be offset onto the virtual surface by the network so that the resulted surface will be less sensitive to their weights. Two examples are shown in Fig.~\ref{fig:fit_offset_compre}, in which adding the offset prediction brings more robustness to outliers and avoids the overfitting or the underfitting. More examples of neighboring points before and after being offset can be found in the supplementary materials. 

\subsection{Cascaded Scale Aggregation}

\begin{figure}[]
  \begin{center}
      \includegraphics[width=0.9\linewidth]{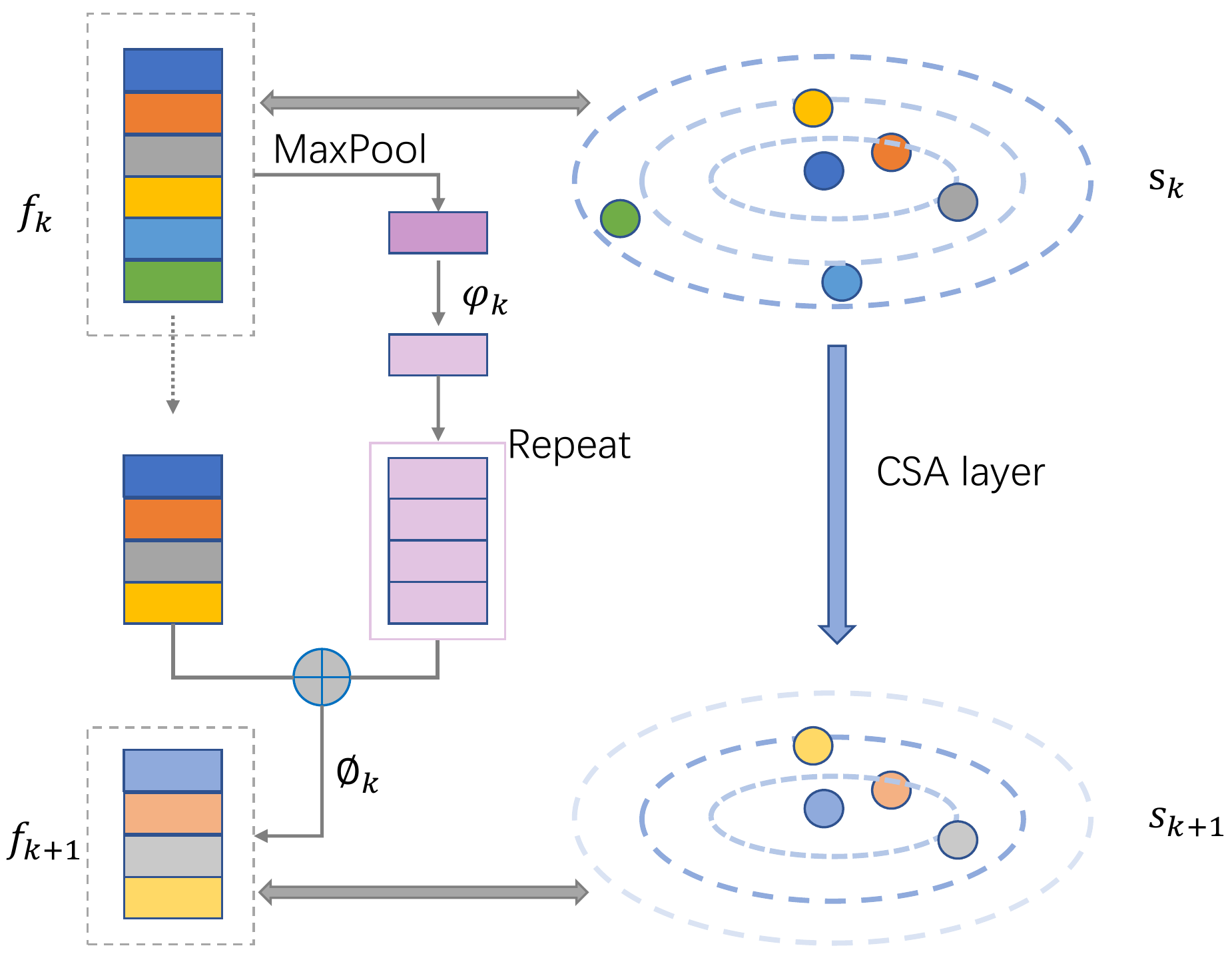}\vspace{-0.45cm}
  \end{center}
  \caption{The structure of a CSA layer, which extracts the features from the large scale by a global pooling and concates the features from the large scale to features of points in a small scale as the output.}
  \label{fig:scale}
  \vspace{-0.2cm}
\end{figure}

In order to extract features for accurate weight and offset prediction, we propose a novel layer called Cascaded Scale Aggregation (CSA) layer. As shown in Fig.~\ref{fig:scale}, the CSA layer is in charge of extracting features from different neighborhood sizes (called scales). 
For a point $p$, we define a scale of this point by an integer $s$, which is represented by a point set $N_s$ including $s$-nearest points to the point.
A CSA layer takes two scales ($s_k$, $s_{k+1}$) as input, where $s_{k+1}<s_k$ so that $N_{s_{k+1}}\subset N_{s_{k}}$.
It extracts a feature $f_{k+1,i}$ on a point $p_i \in N_{s_{k+1}}$ by
\begin{equation}
    f_{k+1,i}=\phi_k (\varphi_k({\rm MaxPool}\{f_{k,j}| p_j \in N_{s_k} \}),f_{k,i})
\end{equation}
where both $\phi_k$ and $\varphi_k$ are Multi-layer Perceptrons (MLP), $f_{k+1,i}$ is the output feature on the point $p_i$ at the scale $s_{k+1}$, $f_{k,i}$ is the feature of the same point at the scale $s_{k}$ and $\{f_{k,j}|p_j \in N_{s_k}\}$ is the set of all features of points in $N_{s_{k}}$. Note that, since $N_{s_{k+1}}\subset N_{s_{k}}$, all points in $N_{s_{k+1}}$ will also be in the $N_{s_{k}}$. 

The CSA layer uses features from a larger scale to help the feature extraction at the current scale while the final fitting only uses the smallest scale. The large scale provides more information of the underlying surfaces while the small scale includes the most relevant points for the surface fitting.
Thus, the proposed CSA layers enjoy the benefits from both the large scale and the small scale, which results in an accurate surface fitting.

\subsection{Implementation details}
\begin{figure*}
\begin{center}
\includegraphics[width=0.9\linewidth]{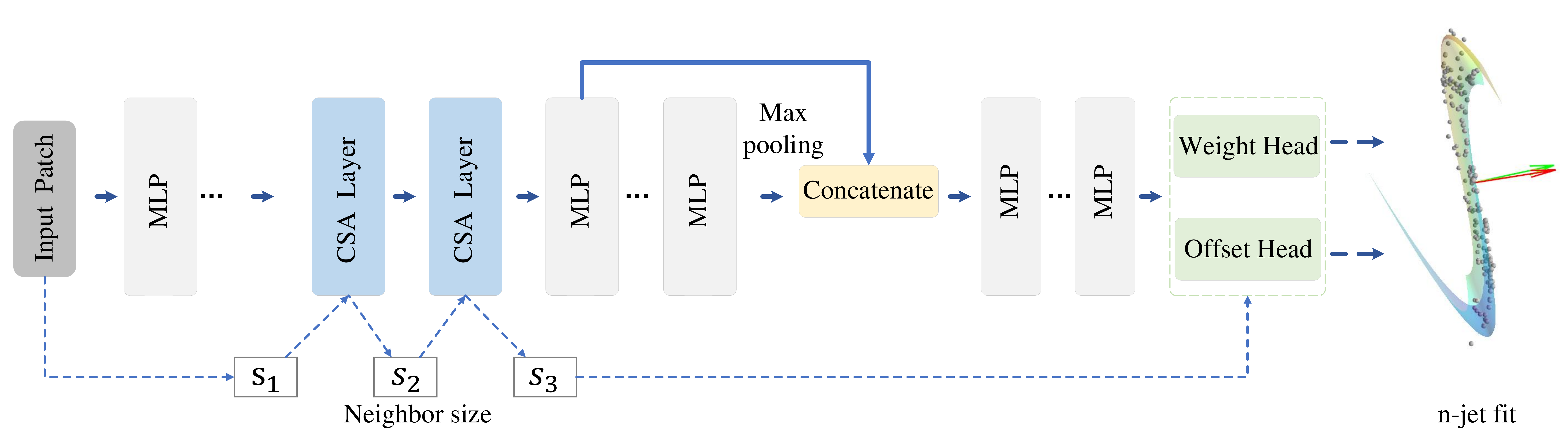}\vspace{-0.5cm}
\end{center}
   \caption{The architecture of AdaFit for normal estimation. AdaFit utilizes CSA layers to extract features from different scales and uses the offset points of the smallest scale with weights to fit a surface. Then the normal vector can be calculated from the surface.}
   
\label{fig:structure}
\end{figure*}

Based on the CSA layers, we design a network called AdaFit which simultaneously predicts point-wise offsets and weights. As shown in Fig.~\ref{fig:structure}, AdaFit mainly consists of MLPs and CSA layers for feature extraction on every point. Then, two heads are used to regress the weights and the offsets.

\textbf{Loss}.To train our network, we follow the same loss as proposed in DeepFit~\cite{ben2020deepfit} which minimizes $|N_{gt}\times N|$ between the predicted normal $N$ and the ground truth normal $N_{gt}$. Meanwhile, we also adopt the neighborhood consistency loss and transformation regularization loss used in DeepFit. Please refer to~\cite{ben2020deepfit} for more details.

\textbf{Parameter setting}.
The polynomial order $n$ for the surface fitting is 3 by default.
For each query point, we use KNN to pick the nearest K-point. 
The initial neighborhood size is 700. AdaFit consists of two CSA layers with the diminishing neighborhood size from 700 to 350, 175.

%% file: 04_experiments.tex
\section{Experiment}

\begin{table*}[!]
\begin{center}
\resizebox{\linewidth}{!}{
  \begin{tabular}{lcccccccl}
  \toprule[1pt]
Aug.                       & AdaFit &  \makecell[c]{DeepFit \\~\cite{ben2020deepfit}} & \makecell[c]{Denoising+\\DeepFit~\cite{ben2020deepfit}} & \makecell[c]{Lenssen \\\etal~\cite{lenssen2020deep}} & \makecell[c]{Nesti-Net\\~\cite{ben2019nesti}} & \makecell[c]{PCPNet\\~\cite{guerrero2018pcpnet}}& PCA~\cite{hoppe1992surface} & Jet~\cite{cazals2005estimating}   \\ \hline
No noise& \textbf{5.19} & 6.51  & 8.48 & 6.72           & 6.99      & 9.66   & 12.29 & 12.23 \\
Noise ($\sigma = 0.00125$) & \textbf{9.05}& 9.21 & 10.38    & 9.95           & 10.11     & 11.46  & 12.87& 12.84 \\
Noise ($\sigma = 0.006$)   & \textbf{16.44}& 16.72 & 16.79  & 17.18& 17.63     & 18.26  & 18.38 & 18.33 \\
Noise ($\sigma = 0.012$)   & \textbf{21.94}& 23.12 & 22.18 & 21.96& 22.28& 22.8& 27.5& 27.68 \\
Varing Density(Strips)& \textbf{6.01}& 7.92 & 9.62 & 7.73& 8.47& 11.74  & 13.66& 13.39 \\
Varing Density(gradients)  & \textbf{5.90}& 7.31 & 9.37   & 7.51& 9.00      & 13.42  & 12.81& 13.13 \\ \hline
Average& \textbf{10.76}& 11.8  & 12.8  & 11.84          & 12.41     & 14.56  & 16.25                   & 16.29 \\ \bottomrule[1pt]
\end{tabular}
}
\vspace{-0.4cm}
\end{center}
\caption{Normal RMSE of AdaFit and baseline methods on the PCPNet dataset.}
\label{table-RMSE} 
 \vspace{-0.15cm}
\end{table*}

\subsection{PCPNet dataset}

We follow exactly the same experimental setup as~\cite{guerrero2018pcpnet} including train-test split, training data augmentation and adding noise or density variations on testing data. 
We use the Adam~\cite{kingma2015adam} optimizer and a learning rate of $5 \times 10^{-4}$ for the training with a batch size of 256. AdaFit is trained with 600 epochs on a 2080Ti GPU.

\textbf{Baselines}. We consider three types of baseline methods: 1) the traditional normal estimation methods PCA~\cite{hoppe1992surface} and jet~\cite{cazals2005estimating}; 2) the learning-based surface fitting methods Lenssen \etal~\cite{lenssen2020deep}, DeepFit~\cite{ben2020deepfit}; 3) the learning-based normal regression methods PCPNet~\cite{guerrero2018pcpnet} and Nesti-Net~\cite{ben2019nesti}.

\textbf{Metrics}. We use the angle RMSE between the predicted normal and the ground truth as our main metrics for evaluation. We also draw the AUC curve of normal errors to show the error distribution.

\begin{figure}[]
\begin{center}
\includegraphics[width=0.9\linewidth]{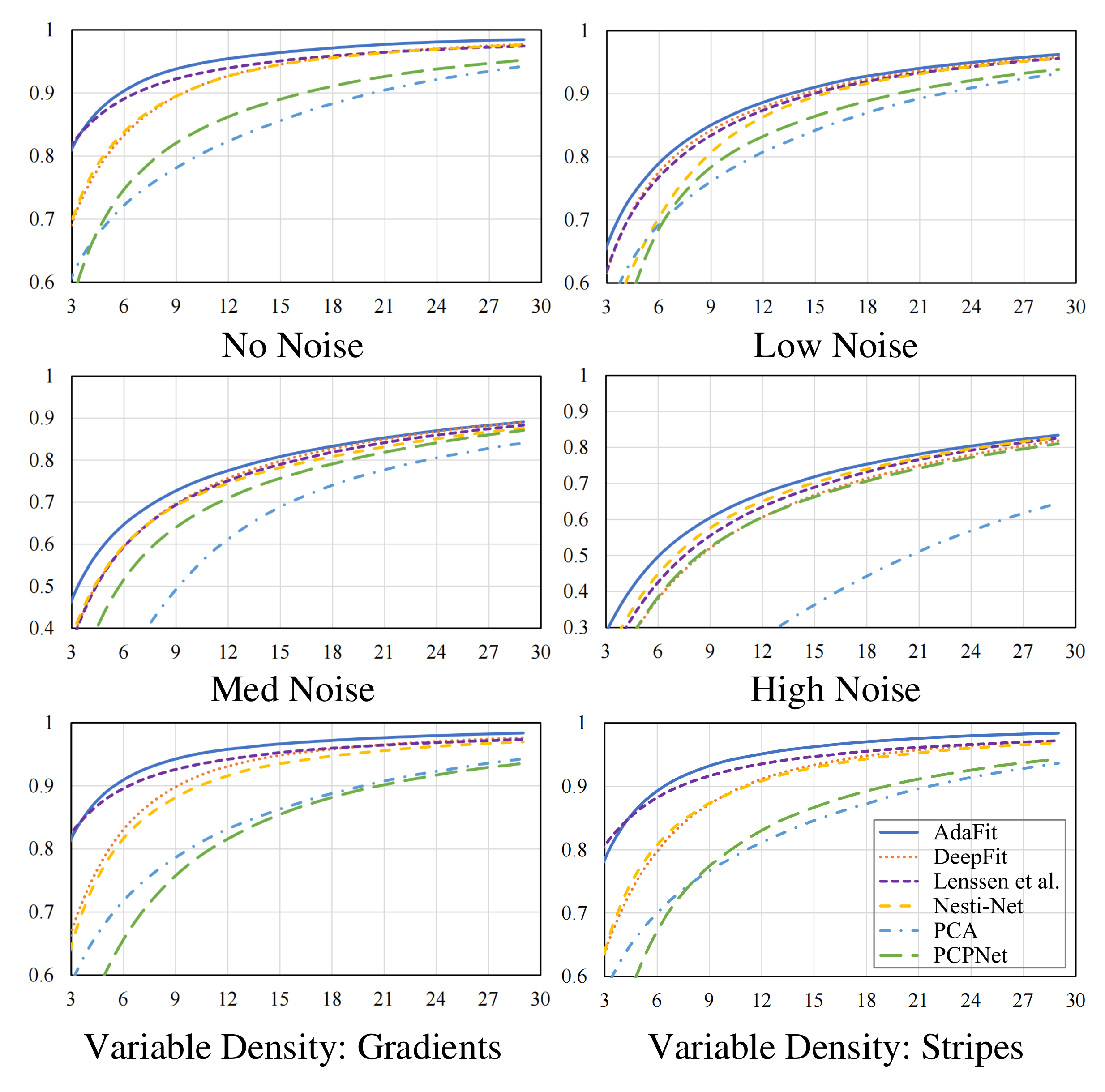}\vspace{-0.5cm}
\end{center}
   \caption{Normal error AUC curves of AdaFit and baseline methods. X-axis shows the threshold in degree and Y-axis shows the ratio of correct estimated normals under a given threshold.}
\label{fig:PCG}
\vspace{-0.4cm}
\end{figure}

\textbf{Quantitative results}. The RMSE of AdaFit and baseline methods are shown in Table~\ref{table-RMSE} and the AUC of all methods are shown in Fig.~\ref{fig:PCG}. The results show that AdaFit outperforms both the traditional methods and learning-based methods in all settings, which demonstrates the effectiveness of using offsets to adjust the point set. Especially on the point clouds with density variations, baseline methods may fail to find enough points on sparse regions for the surface fitting while AdaFit use the offset to project points to the neighboring regions for more robust surface fitting. In addition, we adding the results of DeepFit~\cite{ben2020deepfit} with denoising pre-processing~\cite{rakotosaona2020pointcleannet}. Though denoising reduces the noise and finds smooth surfaces, it does not offset the points to the ground truth positions on the surfaces, which still results in a larger RMSE.

\begin{figure}[]
\begin{center}
\includegraphics[width=0.9\linewidth]{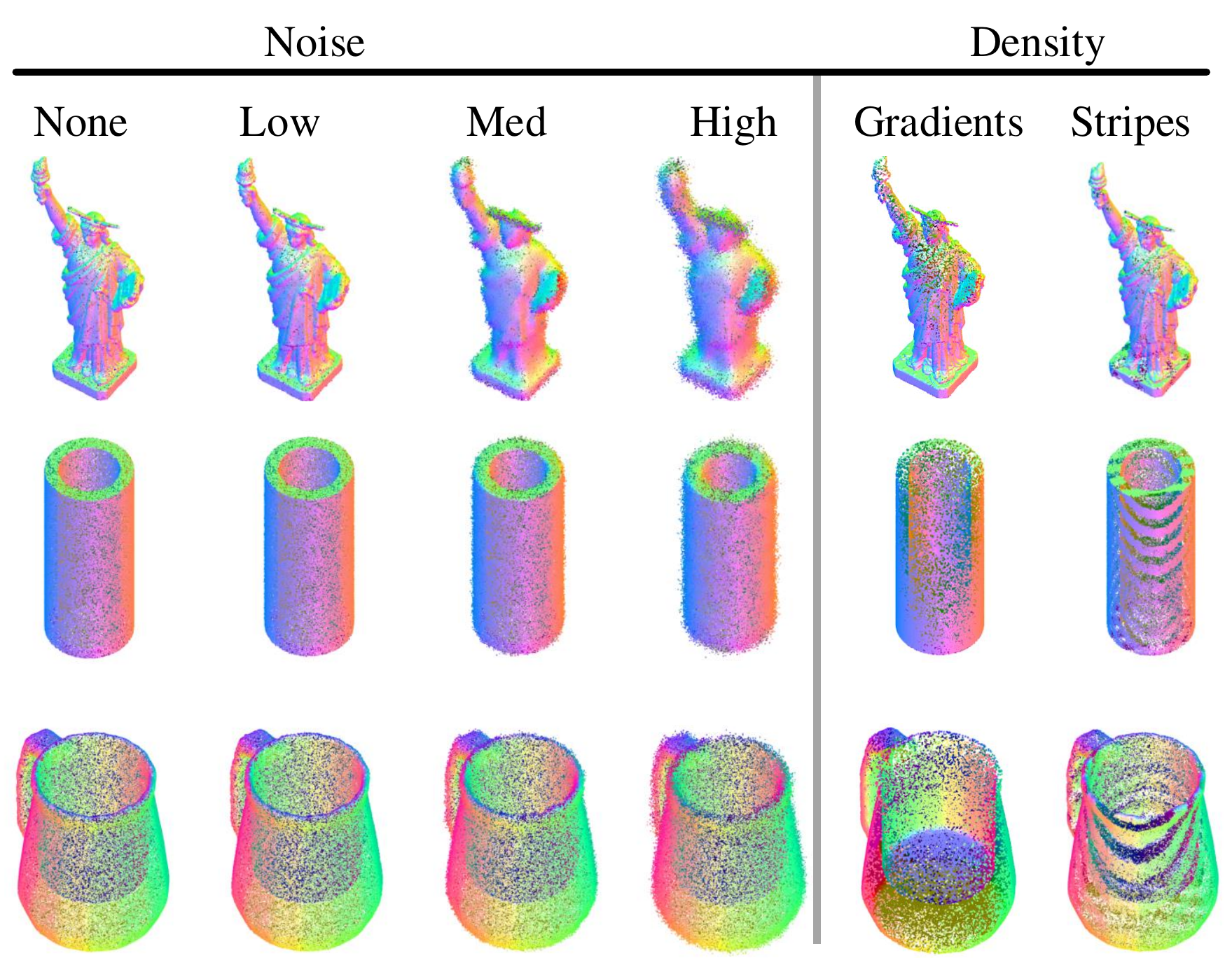}\vspace{-0.3cm}

\end{center}
   \caption{Normals estimated by AdaFit on the PCPNet dataset.}
\label{fig:normal-map}
\vspace{-0.2cm}

\end{figure}

\begin{figure}[]
\begin{center}
\includegraphics[width=0.9\linewidth]{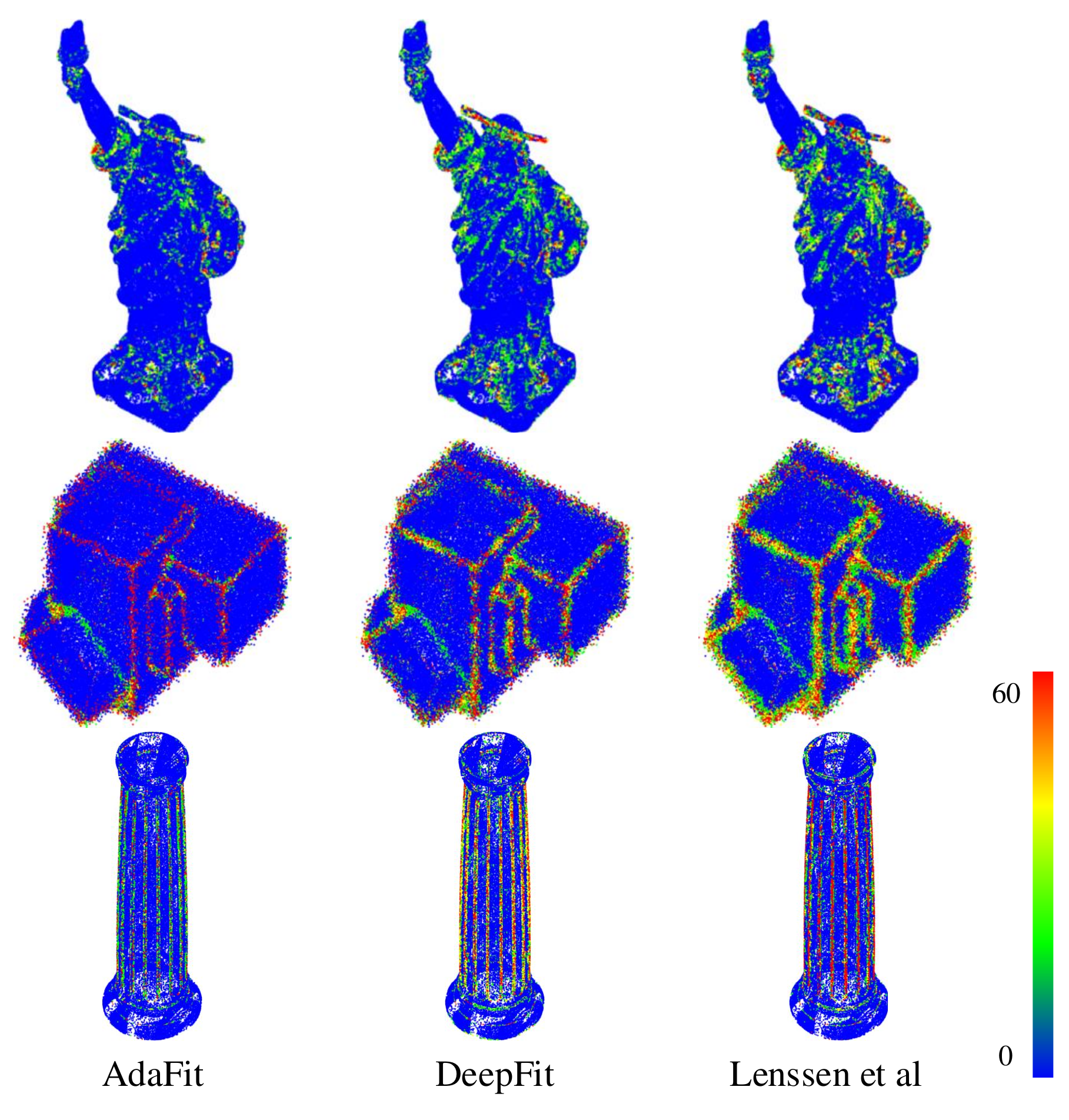}\vspace{-0.5cm}
\end{center}
   \caption{Errors of normal estimation on the PCPNet dataset.}
   
\label{fig:normal-error}
\end{figure}

\textbf{Qualitative results}. Fig.~\ref{fig:normal-map} shows the normals estimated by AdaFit and Fig.~\ref{fig:normal-error} visualizes the angle errors of AdaFit and baseline methods on different shapes. It can be seen that Lenssen \etal~\cite{lenssen2020deep} perform better on the flattened regions while DeepFit~\cite{ben2020deepfit} can better handle curved regions but is not accurate for regions with sharp curves. In contrast, the proposed AdaFit is relatively more robust on all regions and all settings (noise or density variations). Additional detailed analysis on normal estimation of sharp edges can be found in the supplementary materials.

\subsection {Results on real datasets}
\label{Transfer-new}

To test the generalization capability of AdaFit, we directly evaluate AdaFit on real datasets including the indoor SceneNN~\cite{hua2016scenenn} dataset and the outdoor Semantic3D~\cite{hackel2017isprs} dataset. Note that all the methods are only trained on the PCPNet dataset and directly evaluated on these datasets.

\begin{figure}[]
   \begin{center}
      \includegraphics[width=0.9\linewidth]{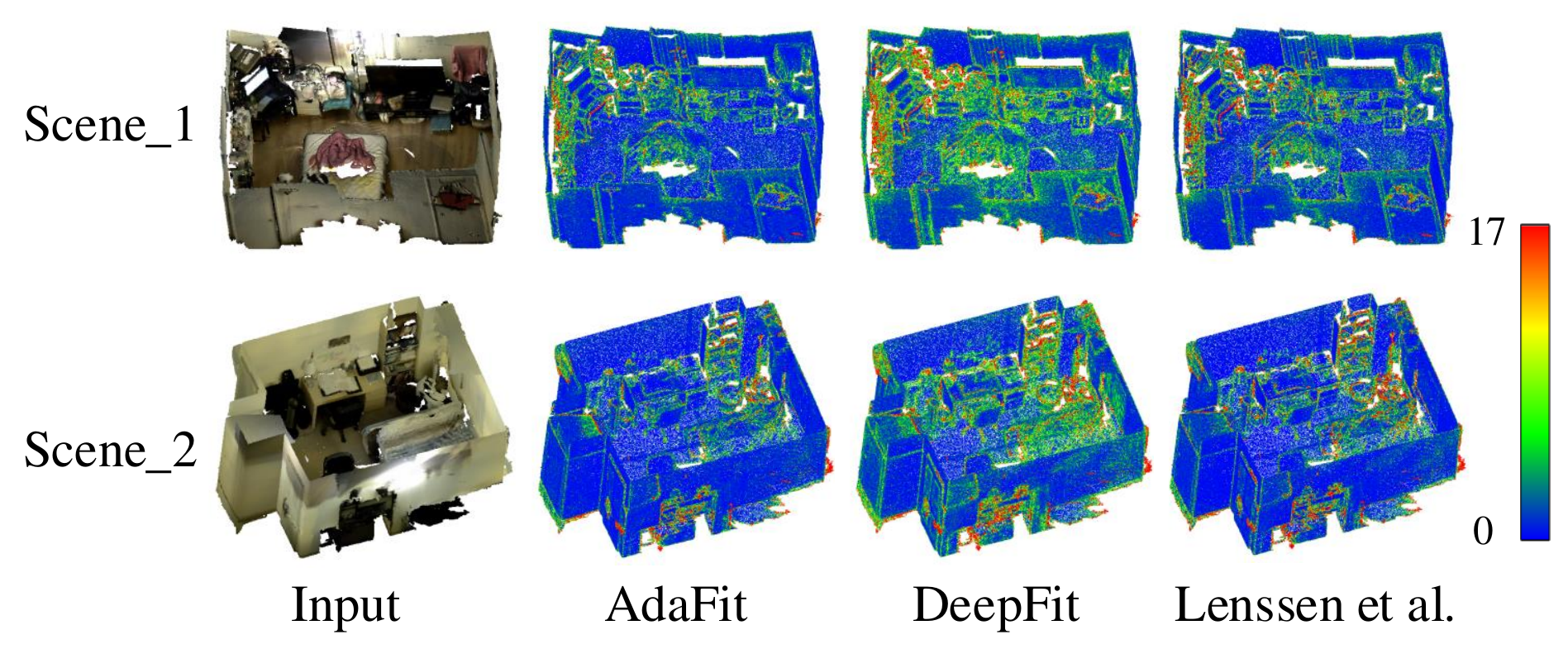}\vspace{-0.4cm}
   \end{center}
   \caption{Error maps of estimated normals on the SceneNN dataset.
   }
   \label{indoor}
\end{figure}

\textbf{The SceneNN dataset}. 
The SceneNN dataset contains more than 100 indoor scenes collected by a depth camera with provided ground-truth reconstructed meshes. We obtain the point clouds by sampling on the reconstructed meshes and compute the ground-truth normal from the meshes. The RMSE of AdaFit and baselines are shown in Table~\ref{RMSE_error_indoor} and the angle error visualizations are shown in Fig.~\ref{indoor}. The results show that AdaFit can predict more accurate normals than baseline methods.

\begin{figure}[]
   \begin{center}
      \includegraphics[width=0.9\linewidth]{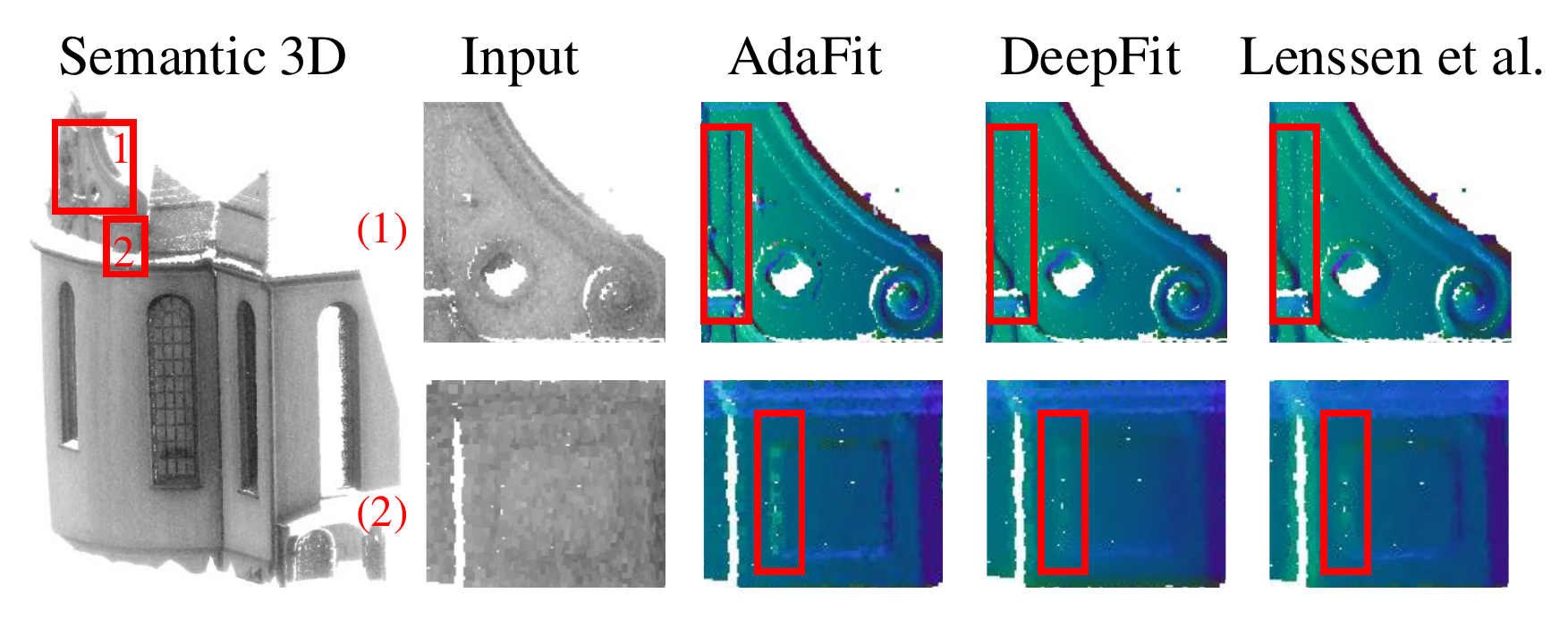}\vspace{-0.5cm}
   \end{center}
   \caption{Estimated normals on the semantic 3D dataset.}
   \label{Semantic3D}
\end{figure}

\textbf{The Semantic3D dataset}. The Semantic3D dataset contains point clouds collected by laser scanners. Since there is no ground truth for normal estimation, we only show qualitative results in Fig.~\ref{Semantic3D}. The results show that AdaFit can find sharp edges on point clouds while other methods oversmooth these normals.

\begin{table}[]
\begin{center}
    \begin{tabular}{cccc}
\toprule[1pt]
& \multicolumn{1}{l}{AdaFit} &\multicolumn{1}{l}{DeepFit~\cite{ben2020deepfit}}&\multicolumn{1}{l}{Lenssn \etal~\cite{lenssen2020deep}} \\ \hline
RMSE & \textbf{16.25}& 18.33& 18.54\\
\bottomrule[1pt]
\end{tabular}
\vspace{-0.2cm}
\end{center}
\caption{Normal RMSE of AdaFit, DeepFit~\cite{ben2020deepfit} and Lenssen \etal~\cite{lenssen2020deep} on the SceneNN dataset.}
\label{RMSE_error_indoor}
\end{table}

\subsection {Ablation study}

\begin{table}[]
\begin{center}
    \setlength{\tabcolsep}{1.5mm}{
        \begin{tabular}{l|cc|cc|cc}
        
        \toprule[1pt]
\textbf{\textbf{  n}} & \multicolumn{2}{c|}{1} & \multicolumn{2}{c|}{2} & \multicolumn{2}{c}{3} \\ \hline
\textbf{Weight} & \checkmark & \checkmark & \checkmark & \checkmark & \checkmark & \checkmark \\
\textbf{Offset} &  & \checkmark &  & \checkmark &  & \checkmark \\
% \hline
No Noise & 8.09 & 5.96 & 10.07 & 6.06 & 7.91 & 6.00 \\
Low Nose & 10.49 & 9.34 & 11.30 & 9.28 & 9.82 & 9.27 \\
Med Noise & 16.59 & 16.56 & 16.59 & 16.48 & 16.36 & 16.48 \\
High Noise & 21.80 & 21.82 & 21.61 & 21.77 & 21.48 & 21.76 \\
Stripes & 9.72 & 7.10 & 12.00 & 7.18 & 9.80 & 7.09 \\
Gradients & 8.54 & 6.67 & 10.83 & 6.80 & 8.59 & 6.72 \\ \hline
Average & 12.54 & 11.24 & 13.73 & 11.26 & 12.33 & 11.22 \\ 
\bottomrule[1pt]

        \end{tabular}
    }
\vspace{-0.4cm}
\end{center}
\caption{Normal RMSE of models with or without offset prediction on the PCPNet dataset.}
\label{Ablation_ADA}

\end{table}

\begin{table}[]
\begin{center}
\setlength{\tabcolsep}{4mm}{
    \begin{tabular}{l|ccc|c}
\toprule[1pt]
\textbf{Scale} & 256 & 500 & 700 & 700 \\
\hline
\textbf{Weight} & \multicolumn{1}{c}{\checkmark} & \checkmark & \multicolumn{1}{c|}{\checkmark} & \multicolumn{1}{c}{\checkmark} \\
\textbf{Offset} & \multicolumn{1}{c}{\checkmark} & \checkmark & \multicolumn{1}{c|}{\checkmark} & \multicolumn{1}{c}{\checkmark} \\
\textbf{CSA} & \multicolumn{1}{l}{} &  & \multicolumn{1}{c|}{} & \multicolumn{1}{c}{\checkmark} \\  
% \hline
No Noise & 5.17 & 5.79 & 6.00 & 5.19 \\
Low Noise & 9.17 & 9.17 & 9.27 & 9.05 \\
Med Noise & 16.71 & 16.47 & 16.48 & 16.44 \\
High Noise & 23.02 & 22.12 & 21.76 & 21.94 \\
Stripes & 6.03 & 6.64 & 7.09 & 6.01 \\
Gradients & 6.00 & 6.30 & 6.72 & 5.90 \\
\hline
Average & 11.02 & 11.08 & 11.22 & 10.76 \\ 
\bottomrule[1pt]
\end{tabular}}
\vspace{-0.4cm}
\end{center}
\caption{Normal RMSE of models with or without CSA layers on the PCPNet dataset.}
\label{Ablation_CSA} 
 \vspace{-0.1cm}
\end{table}

\begin{table}[]
\begin{center}
\setlength{\tabcolsep}{1.5mm}{
    \begin{tabular}{lcccccc}
    
    \toprule[1pt]
          \textbf{Threshold}       & \textbf{0.0}  & \textbf{0.05}   & \textbf{0.10}   & \textbf{0.30}   & \textbf{0.50}   &  \textbf{Offset} \\
          \hline
    No Noise   & 7.91  & 7.35  & 7.41  & 7.94  & 8.41  & 6.00   \\
    Low Noise  & 9.82  & 9.53  & 9.57  & 10.07 & 10.49 & 9.27   \\
    Med Noise  & 16.36 & 16.31 & 16.37 & 16.98 & 18.06 & 16.48  \\
    High Noise & 21.48 & 21.43 & 21.44 & 22.22 & 23.42 & 21.76  \\
    Stripes  & 9.80  & 8.97  & 8.84  & 8.94  & 9.61  & 7.09   \\
    Gradients    & 8.59  & 8.09  & 8.11  & 8.52  & 9.11  & 6.72   \\
    \hline
    Avarage    & 12.33 & 11.95 & 11.96 & 12.44 & 13.18 & 11.22  \\
    
    \bottomrule[1pt]
    \end{tabular}}
    \vspace{-0.4cm}
\end{center}

\caption{Normal RMSE of models using thresholds to truncate points and the model with offset prediction on the PCPNet dataset. The first row shows the truncation thresholds.}
\vspace{-0.4cm}
\label{fig:truncate}
\end{table}

\textbf{Offset prediction}. To demonstrate the effectiveness of the proposed offset prediction, we conduct experiments on the PCPNet dataset using models with or without offset prediction. The backbone network in this experiment is the same as DeepFit without any CSA layer and uses the neighborhood size of 700 points. The results are shown in Table~\ref{Ablation_ADA}, which demonstrates that the offset prediction can effectively improve the accuracy of the predicted normals. Meanwhile, we notice that the performance of the model without offset prediction is sensitive to the polynomial order $n$ while the model with offset prediction has similar performance on different orders.

\textbf{CSA layer}. In Table~\ref{Ablation_CSA}, we conduct experiments on the PCPNet dataset using models with or without CSA layers. The results show that using CSA layers can bring improvements on the normal estimation and reduce the necessity to select a specific neighborhood size (scale).

\textbf{Comparison to truncating weights}. Since outliers may have small predicted weights, a more simple way to prevent these outliers from affecting the estimated normals is to truncate points with a predefined threshold. In Table~\ref{fig:truncate}, we compare the model using proposed offset predictions with the model that truncates points with small weights. All models take 700 neighboring points as inputs. The results show that truncating indeed slightly improves the results but is still inferior to the model with offset prediction.

%% file: 05_applications.tex
\section{Application}
\subsection {Surface reconstruction}
\begin{figure}[]
\begin{center}
\includegraphics[width=0.95\linewidth]{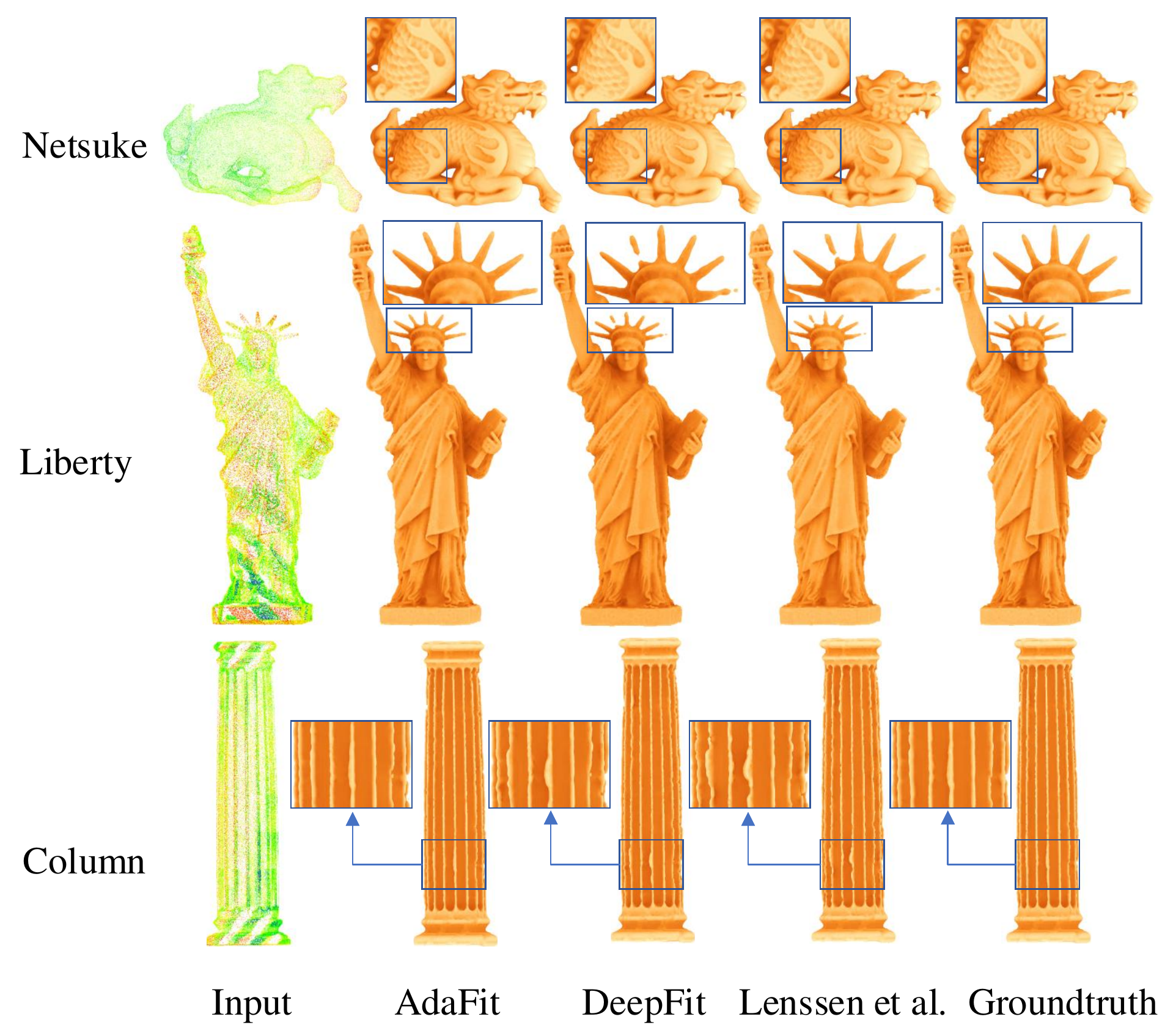}\vspace{-0.45cm}
\end{center}
   \caption{The comparison of the Poisson surface reconstruction using the estimated normals from different methods.
   }
  
\label{fig:poisson}
\vspace{-0.5cm}
\end{figure}

Accurate normals can help the Poisson reconstruction~\cite{kazhdan2006poisson} to reconstruct a more high-quality and complete surface from point clouds. In Fig.~\ref{fig:poisson}, we show surfaces reconstructed by normals estimated with different methods and the corresponding distance RMSE of reconstructed surface is shown in Table~\ref{tab:poisson}. The results show that the reconstructed surface using the normals from AdaFit is more accurate and complete than baseline methods.

\begin{table}[]
\begin{center}
    \begin{tabular}{lccc}
\toprule[1pt]
& \multicolumn{1}{l}{AdaFit} &\multicolumn{1}{l}{DeepFit~\cite{ben2020deepfit}}&\multicolumn{1}{l}{Lenssn \etal~\cite{lenssen2020deep}} \\ \hline
 Netsuke & \textbf{0.00373}& 0.00821& 0.00677\\
 Liberty & \textbf{0.00065} & 0.00100 & 0.00106\\
 Column & \textbf{0.01148} &  0.02060 & 0.03420\\
 \hline
 Average & \textbf{0.00529} & 0.00994 & 0.01401\\
\bottomrule[1pt]
\end{tabular}
\vspace{-0.4cm}
\end{center}
% \vspace{5pt}
\caption{The comparison of the RMSE surface distance error for surface reconstruction of our method and DeepFit~\cite{ben2020deepfit}, Lenssen \etal~\cite{lenssen2020deep}}
 \vspace{-0.3cm}
\label{tab:poisson}
\end{table}

\subsection {Denoising}
\begin{figure}[] 
   \begin{center}
      \includegraphics[width=0.95\linewidth]{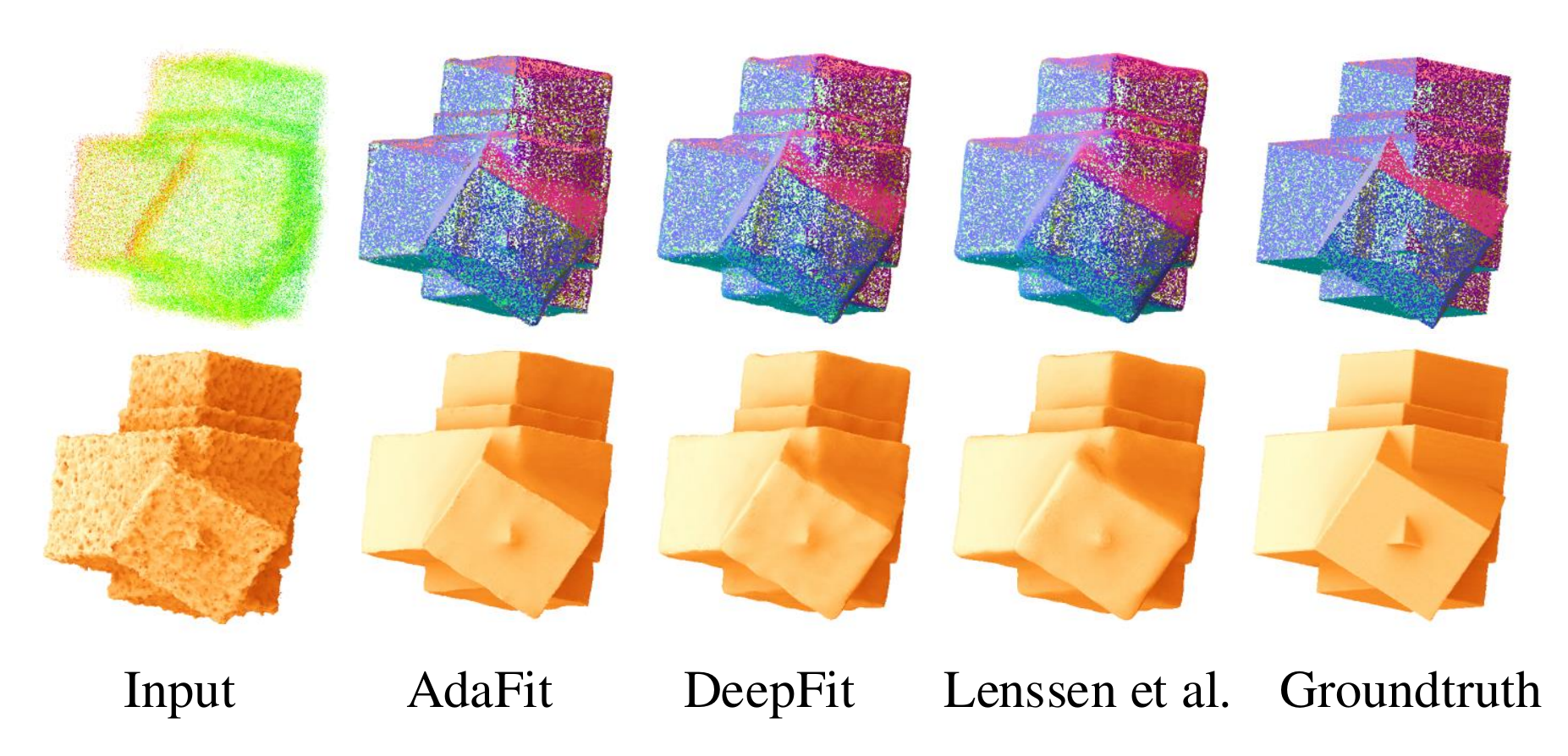}\vspace{-0.5cm}
   \end{center}
   \caption{Qualitative results of point cloud denoising. The first row shows the denoised point clouds while the second row shows the corresponding reconstructed surfaces.}
   \label{fig:denoise}
   \vspace{-0.4cm}
\end{figure}

To demonstrate the application of AdaFit in point cloud denoising, we adopt the normal-based denoising method proposed in~\cite{lu2020low}. The qualitative results of denoised point clouds and their corresponding reconstructed surfaces are shown in Fig.~\ref{fig:denoise}, which indicates that using the normal estimated by AdaFit for denoising produces a smooth surface in flattened regions while still keeping sharp features at edges.

%% file: 06_conclusion.tex
\section{Conclusion}
In this paper, we presented AdaFit for the normal estimation on point clouds. We provided a comprehensive analysis on current weighted least square surface fitting for normal estimation and found two inherent problems of this kind of method. To solve these problems, we proposed the offset prediction on current network and used a novel CSA layer for feature extraction. AdaFit achieved state-of-the-art performance on the PCPNet dataset and strong generalization capability to real-world datasets. Furthermore, we also demonstrated the effectiveness of predicted normals of AdaFit on several downstreams tasks.

%% file: 08_Acknowledgement.tex
\section{Acknowledgement}

This work was supported by: The National Science Fund for Distinguished Young Scholars of China, grant number 41725005; The National Natural Science Foundation Project (No.41901403); the Fundamental Research Funds for the Central Universities (No.2042020kf0015); The Technical Cooperation Agreement between Wuhan University and Huawei Space Information Technology Innovation Laboratory (No.YBN2018095106)

% 1) The National Key Research and Development Program of China (No.2018YFB2100503); 2) The Technical Cooperation Agreement between Wuhan University and Huawei Space Information Technology Innovation Laboratory (No.YBN2018095106); 3) The Fundamental Research Funds for the Central Universities (No.2042020kf0015); 4) The National Natural Science Foundation Project (No.41901403) 